# Exploiting generalisation symmetries in accuracy-based learning classifier systems: An initial study

Larry Bull

**Abstract.** Modern learning classifier systems typically exploit a niched genetic algorithm to facilitate rule discovery. When used for reinforcement learning, such rules represent generalisations over the state-action-reward space. Whilst encouraging maximal generality, the niching can potentially hinder the formation of generalisations in the state space which are symmetrical, or very similar, over different actions. This paper introduces the use of rules which contain multiple actions, maintaining accuracy and reward metrics for each action. It is shown that problem symmetries can be exploited, improving performance, whilst not degrading performance when symmetries are reduced.

## 1 INTRODUCTION

Learning Classifier Systems (LCS) [Holland, 1976] are rule-based systems, where the rules are usually in the traditional production system form of "IF condition THEN assertion". An evolutionary algorithm and/or other heuristics are used to search the space of possible rules, whilst another learning process is used to assign utility to existing rules, thereby guiding the search for better rules. LCS are typically used as a form of reinforcement learner, although variants also exist for supervised [Bernadó Mansilla & Garrell, 2003], unsupervised [Tammee et al., 2007] and function [Wilson, 2002] learning. Almost twenty years ago, Stewart Wilson introduced a form of LCS in which rule utility is calculated solely by the accuracy of the predicted consequences of rule assertions/actions – the "eXtended Classifier System" (XCS) [Wilson, 1995]. Importantly, XCS makes a clear connection between LCS and modern reinforcement learning (see [Sutton & Barto, 1998]): XCS uses a genetic algorithm (GA) [Holland, 1975] to discover regularities in the problem thereby enabling generalisations over the complete state-action-reward space. It has been found able to solve a number of well-known problems optimally (e.g., see [Butz, 2006]). Modern LCS, primarily XCS and its derivatives, have been applied to a number of real-world problems (e.g., see [Bull, 2004]), particularly data mining (e.g., see [Bull et al., 2008]), to great effect. Formal understanding of modern LCS has also increased in recent years (e.g., see [Bull & Kovacs, 2005]).

XCS uses a niched GA, that is, it runs the GA over rules which are concurrently active. Initially, following [Booker, 1985] (see also [Fogarty, 1994]), the GA was run in the match set [M], i.e., the subset of rules whose condition matches the current state. The primary motivation for restricting the GA in this way is to avoid the recombination of rule conditions which generalise over very different areas of the problem space. Wilson [1998] later increased the niching to action sets [A], i.e., the subset of [M] whose action matches the chosen output of the system. Wilson correctly highlighted that for tasks with asymmetrical generalisations per action, the GA would still have the potential to unhelpfully recombine rules working over different sub-regions of the input space unless it is moved to [A]. Using two simple benchmark tasks, he didn't show significant changes in performance but did show a decrease in the number of unique rules maintained when some asymmetry existed from the use in [A]. Modern XCS uses the [A] form of GA, which has been studied formally in various ways (e.g., see [Bull, 2002; 2005][Butz et al., 2004][Butz et al., 2007]). It can be noted that the first LCS maintained separate GA populations per action [Holland & Reitman, 1978] (see [Wilson, 1985] for a similar scheme).

The degree of symmetry within the state-action-reward space across all problems is a continuum. As noted, running the GA in niches of concurrently active rules identifies those whose conditions overlap in the problem space. However, using the GA in [A] means that any common structure in the problem space discovered by a rule with one action must wait to be shared through the appropriate mutation of its action. Otherwise it must be rediscovered by the GA for rules with another action(s). As the degree of symmetry in the problem increases, so the potentially negative effect of using the GA in [A] on the search process increases.

This paper proposes a change in the standard rule structure to address the issue and demonstrates it using a slightly simplified version of XCS, termed YCS [Bull, 2005].

## 2 YCS: A SIMPLE ACCURACY-BASED LCS

YCS is without internal memory, the rule-base consists of a number ($P$) of condition-action rules in which the condition is a string of characters from the traditional ternary alphabet {0,1,#} and the action is represented by a binary string. Associated with each rule is a predicted reward value ($r$), a scalar which indicates the error ($\varepsilon$) in the rule's predicted reward and an estimate of the average size of the niches in which that rule participates ($\sigma$). The initial random population has these parameters initialized, somewhat arbitrarily, to 10.

On receipt of an input message, the rule-base is scanned, and any rule whose condition matches the message at each position is tagged as a member of the current match set [M]. An action is then chosen from those proposed by the members of the match set and all rules proposing the selected action form an action set [A]. A version of XCS's explore/exploit action selection scheme will be used here. That is, on one cycle an action is chosen at random and on the following the action with the highest average fitness-weighted reward is chosen deterministically.

---

Dept. of Computer Science & Creative Tech., UWE BS16 1QY, UK.
Email: larry.bull@uwe.ac.uk.

The simplest case of immediate reward $R$ is considered here. Reinforcement in YCS consists of updating the error, the niche size estimate and then the reward estimate of each member of the current [A] using the Widrow-Hoff delta rule with learning rate $\beta$:

$$\varepsilon_j \leftarrow \varepsilon_j + \beta( |R - r_j| - \varepsilon_j ) \quad (1)$$

$$r_j \leftarrow r_j + \beta( R - r_j ) \quad (2)$$

$$\sigma_j \leftarrow \sigma_j + \beta( |[A]| - \sigma_j ) \quad (3)$$

The original YCS employs two discovery mechanisms, a panmictic (standard global) GA and a covering operator. On each time-step there is a probability $g$ of GA invocation. The GA uses roulette wheel selection to determine two parent rules based on the inverse of their error:

$$f_j = ( 1 / (\varepsilon_j^v + 1) ) \quad (4)$$

Here the exponent $v$ enables control of the fitness pressure within the system by facilitating tuneable fitness separation under fitness proportionate selection (see [Bull, 2005] for discussions). Offspring are produced via mutation (probability $\mu$) and crossover (single point with probability $\chi$), inheriting the parents' parameter values or their average if crossover is invoked. Replacement of existing members of the rulebase uses roulette wheel selection based on estimated niche size. If no rules match on a given time step, then a covering operator is used which creates a rule with the message as its condition (augmented with wildcards at the rate $p_\#$) and a random action, which then replaces an existing member of the rulebase in the usual way. Parameter updating and the GA are not used on exploit trials.

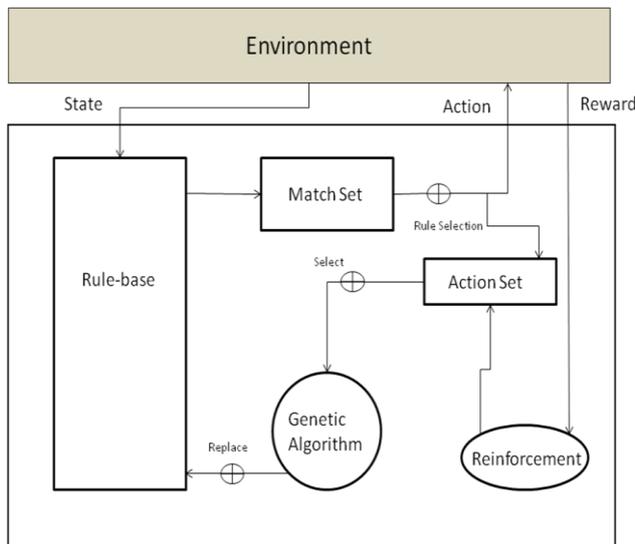

**Figure 1**: Schematic of YCS as used here.

The niche GA mechanism used here is XCS's time-based approach under which each rule maintains a time-stamp of the last system cycle upon which it was part of a GA (a development of [Booker, 1989]). The GA is applied within the current action set [A] when the average number of system cycles since the last GA in the set is over a threshold $\theta_{GA}$. If this condition is met, the GA time-stamp of each rule is set to the current system time, two parents are chosen according to their fitness using standard roulette-wheel selection, and their offspring are potentially crossed and mutated, before being inserted into the rule-base as described above.

YCS is therefore a simple accuracy-based LCS which captures the fundamental characteristics of XCS: "[E]ach classifier maintains a prediction of expected payoff, but the classifier's fitness is *not* given by the prediction. Instead the fitness is a separate number based on an inverse function of the classifier's average prediction error" [Wilson, 1995] and a "classifier's deletion probability is set proportional to the [niche] size estimate, which tends to make all [niches] have about the same size, so that classifier resources are allocated more or less equally to all niches" [ibid]. However, YCS does not include a number of other mechanisms within XCS, such as niche-based fitness sharing, which are known to have beneficial effects in some domains (see [Butz et al., 2004]).

The pressure within XCS and its derivatives to evolve maximally general rules over the problem space comes from the triggered niche GA. Selection for reproduction is based upon the accuracy of prediction, as described. Thus within a niche, accurate rules are more likely to be selected. However, more general rules participate in more niches as they match more inputs. Rules which are both general and accurate therefore typically reproduce the most: the more general and accurate, the more a rule is likely to be selected. Any rule which is less general but equally accurate will have fewer chances to reproduce. Any rule which is over general will have more chances to reproduce but a lower accuracy (see [Butz et al., 2004] for detailed analysis).

Under the new rule representation scheme introduced here each rule consists of a single condition and each possible action. Associated with each action are the two parameters updated according to equations 1 and 2:

Traditional rule – condition: action: reward: error: niche

New rule –     condition: action1: reward: error: niche
                        action2: reward: error
                        action3: reward: error
                        …
                        action$N$: reward: error

All other processing remains the same as described. In this way, any symmetry is directly exploitable by a single rule whilst still limiting the possibility for recombining rules covering different parts of the problem space since the GA is run in [A], as Wilson [1998] described. Any action which is not correctly associated with the generalisation over the problem space represented by the condition will have a low accuracy and can be ignored in any post processing of rules for knowledge discovery. The generalisation process of modern LCS is implicitly extended to evolve rules which are accurate over as many actions as possible since they will participate in more niches. Note that the niche size estimate can become noisier than in standard YCS/XCS. Similarly, any effects from the potential maintenance of inaccurate generalisations in some niches due to their being accurate in other niches are not explored here. Initial results do not indicate any significant disruption however.

## 3 EXPERIMENTATION

### 3.1 Symmetry

Following [Wilson, 1995], the well-known multiplexer task is used in this paper. These Boolean functions are defined for binary strings of length $l = k + 2^k$ under which the $k$ bits index into the remaining $2^k$ bits, returning the value of the indexed bit. A correct classification results in a payoff of 1000, otherwise 0. For example, in the $k=4$ multiplexer the following traditional rules form one optimal [M] (error and niche size not shown):

1111###############1: 1: 1000
1111###############1: 0: 0

Figure 2 shows the performance of YCS using the new multi-action rule representation on the 20-bit multiplexer ($k=4$) problem with $P=1000$, $p_\#=0.6$, $\mu=0.04$, $\nu=10$, $\chi=0.5$, $\theta_{GA}=25$ and $\beta=0.2$. After [Wilson, 1995], performance, taken here to mean the fraction of correct responses, is shown from exploit trials only, using a 50-point running average, averaged over twenty runs. It can be seen that optimal performance is reached around 60,000 trails. Figure 2 also shows the average specificity, taken here to mean the fraction of non-# bits in a condition, for the LCS. That is, the amount of generalization produced. The maximally general solution to the 20-bit multiplexer has specificity $5/20 = 0.25$ and YCS can be seen to produce rule-bases with an average specificity very close to the optimum. The average error of rules can also be seen to decrease over time.

Figure 3 shows the performance of YCS using the traditional rule representation with the same parameters. As can be seen, optimal performance is not reliably reached in the allowed time. Figure 4 shows the performance of the same system with $P=2000$, with optimality reached around 60,000 trials (matching that of XCS with the same equivalent parameters, e.g., [Butz et al., 2004]). That is, with double the rule-base resource, the GA is able to reliably (re)discover the problem structure in all [A] over the same time period using the traditional rule representation. Hence, in a problem with complete symmetry between [A], the new rule representation presented here significantly improves the efficiency of the GA.

### 3.2 Less Symmetry

To reduce the symmetry in the multiplexer in a simple way, an extra bit can be added. Here an incorrect response becomes sensitive to the value of the extra input bit: if it is set, the reward is 500, otherwise it is 0. That is, using the new rule representation, it is no longer possible for just one rule to use the same generalisation over the input space to accurately predict the reward for each action in a given [M]. The following traditional rules represent one optimal [M]:

1111###############1#: 1: 1000
1111###############11: 0: 500
1111###############10: 0: 0

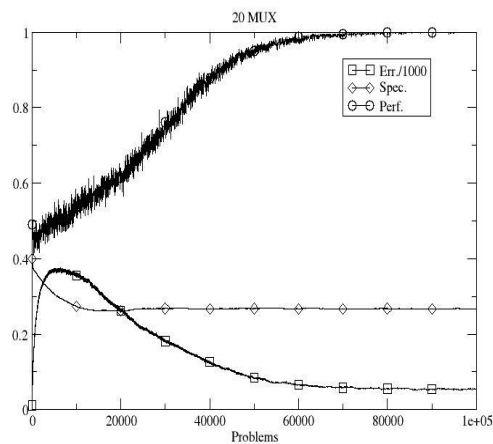

**Figure 2**: Performance of new rule representation.

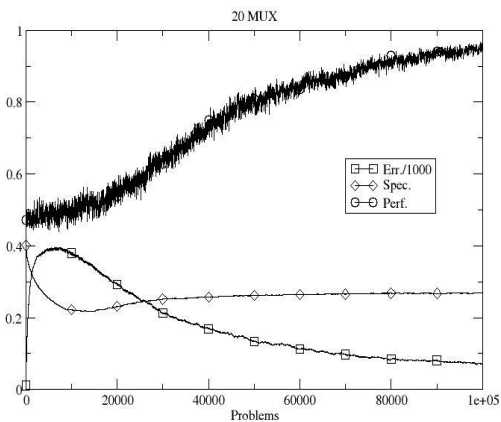

**Figure 3**: Performance of traditional rule representation.

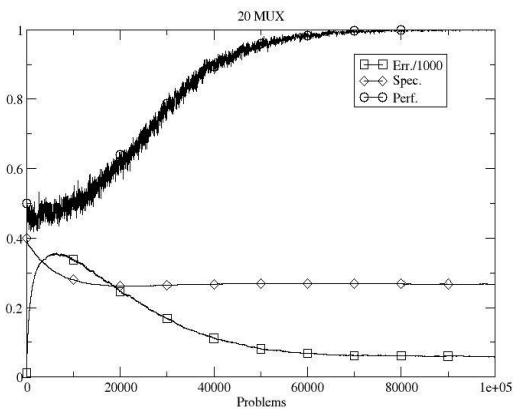

**Figure 4**: As Figure 3 but with larger population size.

Figure 5 shows how YCS is unable to solve the less symmetrical 20-bit multiplexer using the new rule representation with $P=1000$. Figures 6 and 7 show how the performance of YCS with and without the new representation (respectively) is optimal and roughly equal with $P=2000$. Note that the new representation still only requires two rules per [M], as opposed to three in the traditional scheme. However, although there is a slight increase in learning speed with the new scheme, it is not statistically significant (T-test, time taken to reach and maintain optimality over 50 subsequent exploit cycles, $p>0.05$). Figures 8 and 9 show there is significant benefit ($p\leq0.05$) from the new representation when $k=5$, i.e., the harder 37-bit multiplexer ($P=5000$).

### 3.3 Multiple Actions

Multiplexers are binary classification problems. To create a multi-class/multi-action variant in a simple way the case where the data bit is a '1' is altered to require an action equal to the value of the address bits for a correct response. In this way there are $2^k$ possible actions/classes. Under the new format with $k=3$, one optimal [M] could be represented as the single rule:

```
111#######1: 8: 1000
             7: 0
             6: 0
             5: 0
             4: 0
             3: 0
             2: 0
             1: 0
             0: 0
```

Figures 10 and 11 show the performance of YCS with and without the new representation (respectively) with $k=3$ and $P=2000$. As can be seen, both representations are capable of optimal performance with the parameters used but the new representation learns significantly faster (($p\leq0.05$).

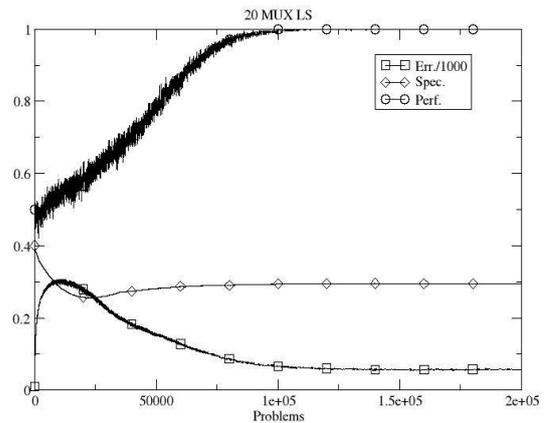

**Figure 6**: As Figure 5 but with larger population size.

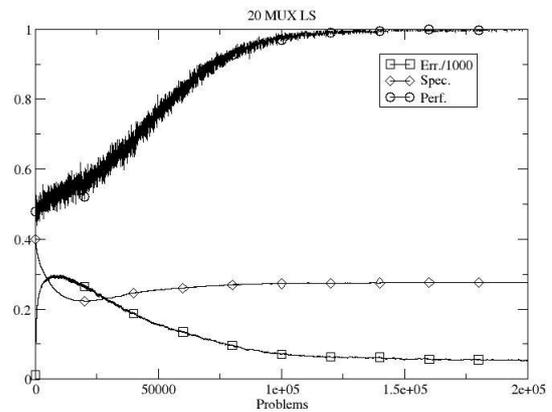

**Figure 7**: Performance of traditional rules on less symmetrical task (vs. Figure 6).

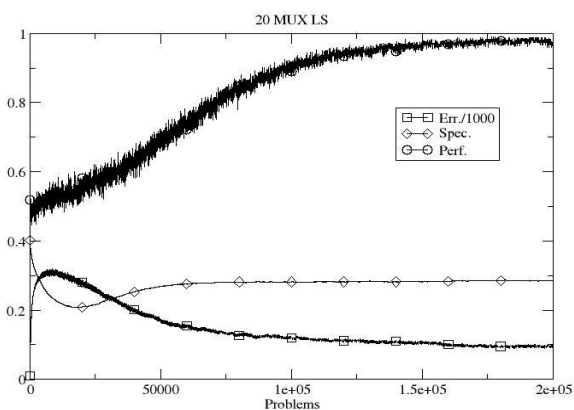

**Figure 5**: Performance of new scheme on less symmetrical task.

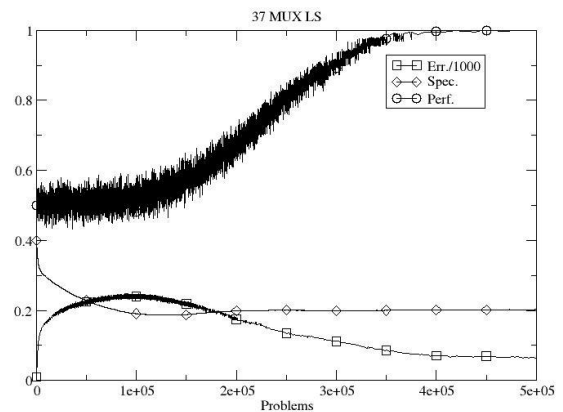

**Figure 8**: Performance of new scheme on less symmetrical multiplexer when $k=5$.

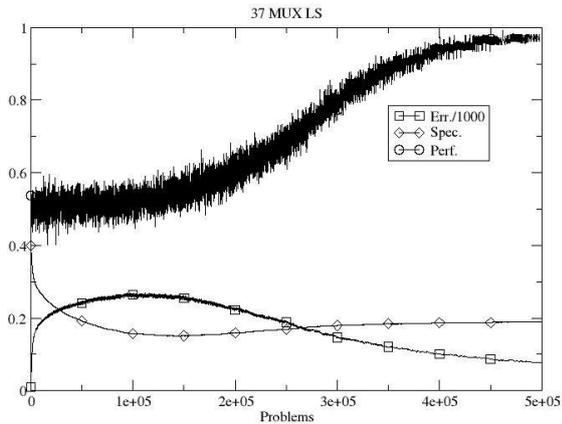

**Figure 9**: Performance of traditional rules on less symmetrical multiplexer when *k*=5.

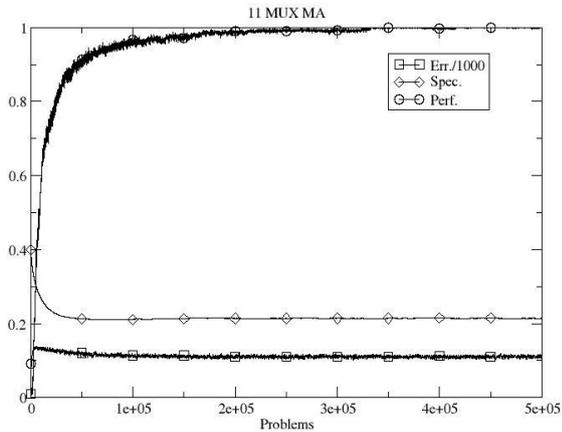

**Figure 10**: Performance of new scheme on multi-action task.

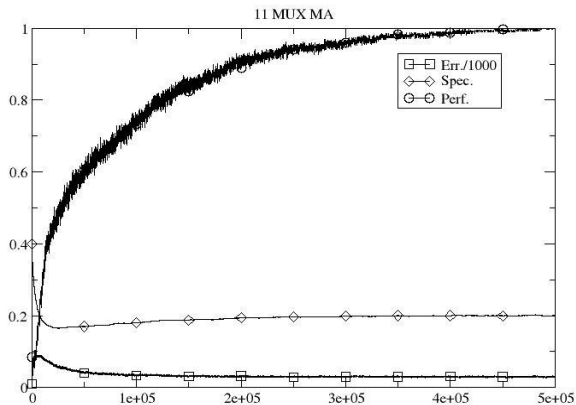

**Figure 11**: Performance of traditional rules on multi-action task.

## 3.4 Imbalance

The frequency of state visitation is rarely close to uniform in most reinforcement learning tasks. For example, in a spatial maze navigation task, those states at or near a goal will typically be visited more often than those states far from a goal. In data mining, real-world data does not typically contain equal examples of all cases of the underlying concept space - known as the class imbalance problem, and often tackled through under/over sampling. This bias of sampling the problem space can cause difficulties in the production of accurate generalisations since over general rules can come to dominate niches due to their frequency of use (updating and reproduction) in more frequently visited states. Orriols-Puig and Bernado Mansilla [2008] introduced a heuristic specifically for (limited to) binary classification tasks which dynamically alters the learning rate ($\beta$) and frequency of GA activity ($\theta_{GA}$) to address the issue in accuracy-based LCS. They show improved learning in both imbalanced multiplexers and well-known data sets.

The new rule representation would appear to have some potential to address the issue of imbalance generally when there is symmetry in the underlying problem space, i.e., both for reinforcement learning and data mining. Since all actions are maintained by all rules, information about all actions is maintained in the population. Whilst over general conditions will quickly emerge for the same reasons as for the traditional representation, later in the search, the use and updating of the correct actions for less frequently visited states will indicate their true value and the GA will (potentially) adjust generalisations appropriately. An imbalanced multiplexer (akin to [Orriols-Puig & Bernado Mansilla, 2008]) can be created by simply introducing a probabilistic bias in sampling action '1' compared to '0'. Figures 12 and 13 show the performance of YCS with and without the new representation (respectively) with *k*=4, *P*=2000 and a bias of 80% (4:1). Exploit cycle testing remains unbiased, as before. As can be seen, the new representation is able to cope with the bias, whereas the equivalent traditional rule representation is not. The same was generally found to be true for various levels of bias, *k*, etc. (not shown).

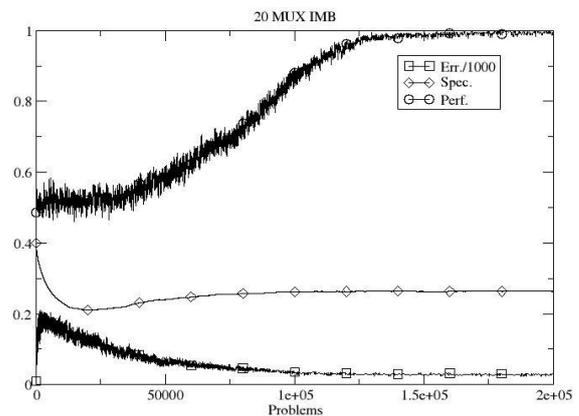

**Figure 12**: Performance of new scheme on the imbalanced task.

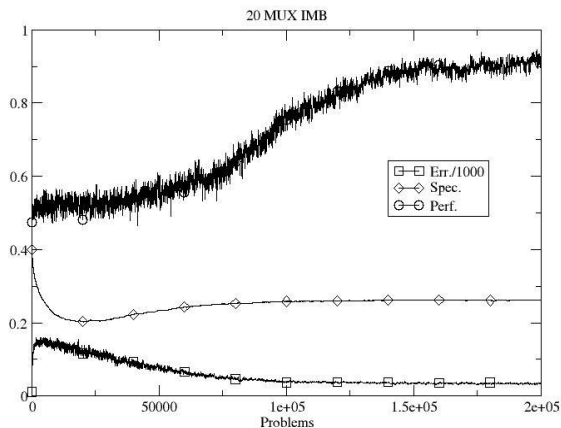

**Figure 13**: Performance of the traditional scheme on the imbalanced task.

## 4 CONCLUSIONS & FUTURE WORK

This paper has proposed the use of rules which contain multiple actions, maintaining accuracy and reward metrics for each action. This somewhat minor alteration appears to provide benefits over the traditional approach in a variety of scenarios. Future work should also consider the new, general rule structure proposed here with more complex representations such as real-valued intervals (e.g., see [Stone & Bull, 2003]) or genetic programming (e.g., see [Preen & Bull, 2013]), together with delayed reward tasks.

Kovacs and Tindale [2013] have recently highlighted issues regarding the niche GA, particularly with respect to overlapping problems. They compare the performance of an accuracy-based LCS with a global GA (see also [Bull, 2005]), a niche GA, and a global GA which uses the calculated selective probabilities of rules under a niche GA. The aim being to avoid the reduced actual selection of accurate, general rules due to overlap within a given niche. Using the 11-bit multiplexer ($k$=3) problem they show a possible slight increase in performance from their new scheme over the niche GA, with the global GA performing worst. Their new scheme shows an increase in the number of unique rules maintained compared to the niche GA and they postulate this increase in rule diversity may explain the suggested difference in performance. This seems likely given the multiplexer does not contain any overlap. Note that Wilson [1994] proposed using both a global and niche GA together "to offset any inbreeding tendency" within niches. Since they used a supervised form of XCS which only maintains the highest reward entries of the state-action-reward map (UCS) [Bernado Mansilla & Garrell, 2003], the exploitation of symmetry does not help to explain their findings. The effect of the new representation in overlapping problems remains to be explored. The related use of multiple conditions per action may be a more appropriate approach.

## REFERENCES

Bernado Mansilla, E. & Garrell, J. (2003) Accuracy-Based Learning Classifier Systems: Models, Analysis and Applications to Classification Tasks. *Evolutionary Computation* 11(3): 209-238.

Booker, L.B. (1985) Improving the Performance of Genetic Algorithms in Classifier Systems. In J.J. Grefenstette (ed) *Proceedings of the First International Conference on Genetic Algorithms and their Applications.* Lawrence Erlbaum Associates, pp80-92.

Booker, L.B. (1989) Triggered Rule Discovery in Classifier Systems. In J. Schaffer (ed) *Proceedings of the Third International Conference on Genetic Algorithms and their Applications.* Morgan Kaufmann, pp265-274.

Bull, L. (2002) On Accuracy-based Fitness. *Soft Computing* 6(3-4): 154-161.

Bull, L. (2004)(ed) *Applications of Learning Classifier Systems.* Springer.

Bull, L. (2005) Two Simple Learning Classifier Systems. In L. Bull & T. Kovacs (eds) *Foundations of Learning Classifier Systems.* Springer, pp63-90.

Bull, L. & Kovacs, T. (2005)(eds) *Foundations of Learning Classifier Systems.* Springer.

Bull, L., Bernado Mansilla, E & Holmes, J. (2008)(eds) *Learning Classifier Systems in Data Mining.* Springer.

Butz, M. (2006) *Rule-based Evolutionary Online Learning Systems.* Springer.

Butz, M., Kovacs, T., Lanzi, P-L & Wilson, S.W. (2004) Toward a Theory of Generalization and Learning in XCS. *IEEE Transactions on Evolutionary Computation* 8(1): 28-46

Butz, M., Goldberg, D., Lanzi, P-L. & Sastry, K. (2007) Problem solution sustenance in XCS: Markov chain analysis of niche support distributions and the impact on computational complexity. *Genetic Programming and Evolvable Machines* 8(1): 5-37

Fogarty, T.C. (1994) Co-evolving Co-operative Populations of Rules in Learning Control Systems. In T.C. Fogarty (ed) *Evolutionary Computing.* Springer, pp195-209.

Holland, J.H. (1975) *Adaptation in Natural and Artificial Systems.* University of Michigan Press.

Holland, J.H. (1976) Adaptation. In R. Rosen & F.M. Snell (eds) *Progress in Theoretical Biology, 4.* Academic Press, pp313-329.

Holland, J.H. & Reitman, J.H. (1978) Cognitive Systems Based in Adaptive Algorithms. In Waterman & Hayes-Roth (eds) *Pattern-directed Inference Systems.* Academic Press.

Kovacs, T. & Tindale, R. (2013) Analysis of the niche genetic algorithm in learning classifier systems. In *Proceedings of the Genetic and Evolutionary Computation Conference.* ACM Press, pp1069-1076.

Orriols-Puig, A. & Bernado Mansilla, E. (2008) Evolutionary Rule-based Systems for Imbalanced Data Sets. Soft Computing 13(3): 213-225.

Preen, R. & Bull, L. (2013) Dynamical Genetic Programming in XCSF. *Evolutionary Computation* 21(3): 361-388.

Stone, C. & Bull, L. (2003) For Real! XCS with Continuous-Valued Inputs. *Evolutionary Computation* 11(3): 299-336

Sutton, R.S. & Barto, A.G. (1998) *Reinforcement Learning.* MIT Press.

Tammee, K., Bull, L. & Ouen, P. (2007) Towards Clustering with XCS. In D. Thierens et al. (eds) *Proceedings of the Genetic and Evolutionary Computation Conference.* ACM Press, pp1854-1860

Wilson, S.W. (1985) Knowledge Growth in an Artificial Animal. J.J. Grefenstette (ed) *Proceedings of the First International Conference on Genetic Algorithms and their Applications.* Lawrence Erlbaum Associates, pp16-23.

Wilson, S.W. (1994) ZCS: A Zeroth-level Classifier System. *Evolutionary Computation* 2(1):1-18.

Wilson, S.W. (1995) Classifier Fitness Based on Accuracy. *Evolutionary Computation* 3(2):149-177.

Wilson, S.W. (1998) Generalization in the XCS Classifier System. In Koza et al. (eds.) *Genetic Programming 1998: Proceedings of the Third Annual Conference.* Morgan Kaufmann, pp322-334.

Wilson, S.W. (2002) Classifiers that Approximate Functions. *Natural Computing* 1(2-3): 211-234.